\pdfoutput=1

\documentclass[11pt]{article}

\usepackage[final]{EMNLP2022}

\usepackage{times}
\usepackage{latexsym}
\usepackage[T1]{fontenc}
\usepackage[utf8]{inputenc}

\usepackage{inconsolata}

\usepackage{listings}
\usepackage{color}

\definecolor{dkgreen}{rgb}{0,0.6,0}
\definecolor{gray}{rgb}{0.5,0.5,0.5}
\definecolor{mauve}{rgb}{0.58,0,0.82}
\definecolor{backcolour}{rgb}{0.95,0.95,0.92}

\usepackage{todonotes}
\usepackage{fancyvrb}
\usepackage{pifont}
\usepackage{scrextend}
\newcommand{\cmark}{\centering \ding{51}}%
\newcommand{\xmark}{ \ding{55}}

\newcommand{\Ucal}{\mathcal{U}}
\newcommand{\Lcal}{\mathcal{L}}
\newcommand{\Vcal}{\mathcal{V}}
\newcommand{\Tcal}{\mathcal{T}}

\newcommand{\bfx}{\mathbf x}

\newcommand{\cage} {\textsc{Cage}}

\newcommand{\spear}{\mbox{\textsc{Spear}}}
\title{\textsc{SPEAR} : Semi-supervised Data Programming in Python}
\author{Guttu Sai Abhishek$^{1}\thanks{~~Authors contributed equally}$,  Harshad Ingole$^{1}\footnotemark[1]$, Parth Laturia$^{1}\footnotemark[1]$, Vineeth Dorna$^{1}\footnotemark[1]$,  \\ \textbf{Ayush Maheshwari$^{1}\footnotemark[1]$, Rishabh Iyer$^{2}$, Ganesh Ramakrishnan$^{1}$}  \\
$^1$Indian Institute of Technology Bombay \\  $^2$The University of Texas at Dallas
}

\begin{document}

\maketitle
      

\begin{abstract}%
We present \spear, an open-source python library for data programming with semi supervision. The package implements several recent data programming approaches including facility to programmatically label and build training data. SPEAR facilitates \textit{weak supervision} in the form of heuristics (or rules) and association of \textit{noisy} labels to the training dataset. These \textit{noisy} labels are aggregated to assign labels to the unlabeled data for downstream tasks. We have implemented several label aggregation approaches  that aggregate the \textit{noisy} labels and then train using the \textit{noisily} labeled set in a cascaded manner. Our implementation also includes other approaches that \textit{jointly} aggregate and train the model for text classification tasks. Thus, in our python package, we integrate several cascade and joint data-programming approaches while also  providing the facility of data programming by letting the user define labeling functions or rules.
The code and tutorial notebooks are available at \url{https://github.com/decile-team/spear}. Further, extensive documentation can be found at \url{https://spear-decile.readthedocs.io/}. Video tutorials demonstrating the usage of our package are available  \href{https://youtube.com/playlist?list=PLW8agt_HvkVnOJoJAqBpaerFb-z-ZlqlP}{here}. We also present some real-world use cases of \spear.

\end{abstract}

\maketitle


\section{Introduction}
Supervised machine learning approaches require large amounts of labeled data to train robust machine learning models. 
For classification tasks such as spam detection, (movie) genre categorization, sequence labelling, and so on, modern machine learning systems rely heavily on human-annotated \textit{gold} labels. Creating labeled data can be a time-consuming and expensive procedure that necessitates a significant amount of human effort.
To reduce dependence on human-annotated labels, various techniques such as semi-supervision, distant supervision, and crowdsourcing have been proposed.
In order to help reduce the subjectivity and drudgery in the labeling process, several recent data programming approaches~\citep{bach2019snorkel,  oishik, awasthi2020learning, spear} have proposed the use of \textit{human-crafted} labelling functions or automatic LFs~\citep{maheshwari2022learning} to \textit{weakly} associate labels with the training data. Users encode supervision in the form of labelling functions (LFs), which assign noisy labels to unlabeled data, reducing dependence on human labeled data. LFs can defined as first-order logic rules as a composition of semantic role attributes \cite{sen2020learning} or syntactic grammar rules \cite{sahay2021rule}.

\begin{figure*}[!t]
    \centering
    \includegraphics[width=0.95\linewidth]{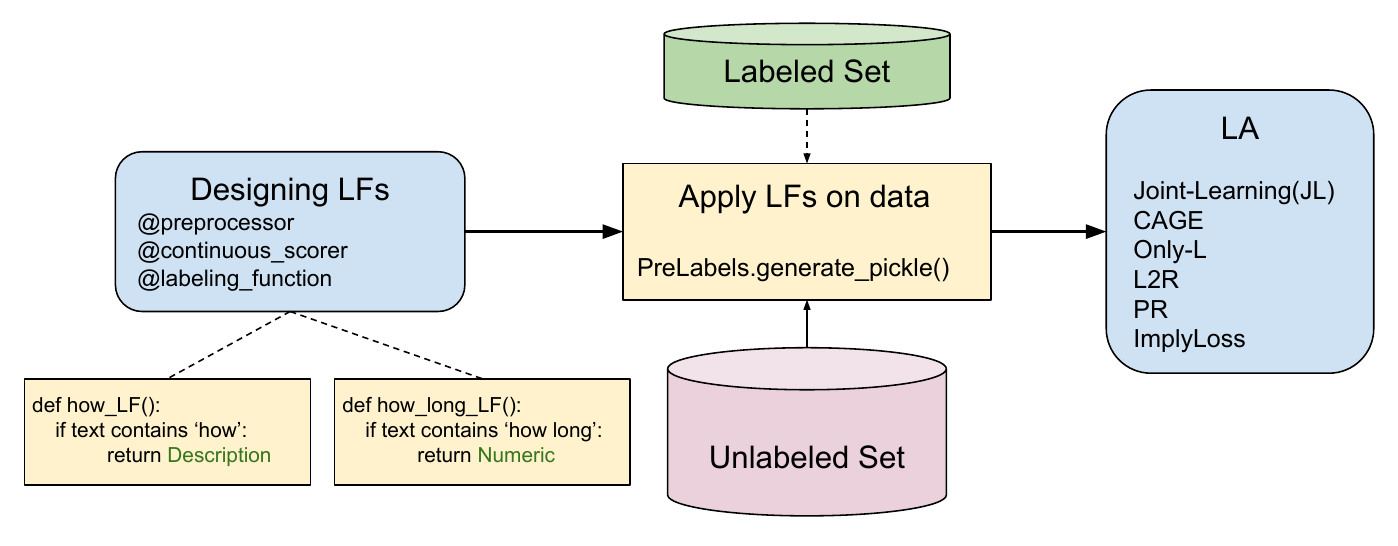}
    \caption{Flow of the \textsc{SPEAR} library.}
    \label{fig:flow}
\end{figure*}

While most data-programming approaches cited above provide their source code in the public domain, a unified package providing access to all data programming approaches is however missing. In this work, we describe \spear, a python package that implements several existing data programming approaches while also providing a platform for integrating and benchmarking newer ones. Inspired by frameworks such as Snorkel~\citep{lison-etal-2021-skweak, ratner2017snorkel, zhang2021wrench} and 
algorithm based labeling in Matlab\footnote{https://www.mathworks.com/help/vision/ug/create-automation-algorithm-for-labeling.html}, we provide a facility for users to define LFs. Further, we develop and integrate several recent data programming models that uses these LFs. We provide many easy-to-use jupyter notebooks and video tutorials for helping new users get quickly started. Though we provide  implementation on 5 text datasets, our package can be easily integrated with vision and speech datasets as well. The users can get started by installing the package using the below command.
\begin{lstlisting}
pip install decile-spear
\end{lstlisting}
In Table \ref{tab:comparisonTable}, we compare our library with other existing packages such as Wrench \cite{zhang2021wrench}, SkWeak\cite{lison-etal-2021-skweak}, Imply Loss \cite{awasthi2020learning}, Snorkel \cite{bach2019snorkel} and Matlab. Wrench \cite{zhang2021wrench} provides facility for semi-supervised and unsupervised label aggregation approaches, however, it does not provide mechanism to find useful subset of unlabeled data and defining continuous LFs. SkWeak \cite{lison-etal-2021-skweak} does not integrate semi-supervised LA approaches in the package. \spear{} addresses the shortcomings of existing packages by providing features such as designing of discrete and continuous LFs, integrating unsupervised and semi-supervised aggregation approaches and facility to choose labeled set using subset selection approaches.

\section{Package Flow}
The \spear\ package consists of three components (and they are applied in the same order): (i) Designing LFs, (ii) applying LFs, and (iii) applying a label aggregator (LA). \\
Initially, the user is expected to declare an \textit{enum} class listing all the class labels. The \textit{enum} class associates the numeric class label with the readable class name.
As part of (i), \spear\ provides the facility for manually creating LFs. LFs can be in the form of regex rules as well. Additionally, we also provide the facility of declaring a \textit{@preprocessor} decorator to use an external library such as \textit{spacy}\footnote{https://spacy.io}, nltk, {\em{etc.}} which can be optionally invoked by the LFs.
Thereafter, as part of (ii), the LFs can be applied on the unlabeled (and labeled) set using an \textit{apply} function that returns a matrix of dimension \#LFs $\times$ \#instances.
The matrix is then provided as input to the selected label aggregator (LA)  in (iii), as shown in Figure\ref{fig:flow}.
We integrate several LA options into \spear. Each LA aggregates multiple noisy labels (obtained from the LFs) to associate a single class label with an instance. Additionally, we have also implemented in \spear, several joint learning approaches that employ semi-supervision and feature information.
The high-level flow of the \spear\ library is presented in Figure \ref{fig:flow}.

    
    \begin{table*}[!h]
    \centering
    \begin{center}
    \begin{tabular}{|p{0.21\textwidth}|p{0.15\textwidth}|p{0.12\textwidth}|p{0.1\textwidth}|p{0.1\textwidth}|p{0.17\textwidth}|}
    \hline
         Package & Designing \& applying LFs & Continuous LFs & Unsup LA & Semi-sup LA & Labeled-data subset selection  \\
         \hline
         Snorkel\cite{ratner2017snorkel}  & \centering \cmark  & \centering \xmark &  \centering \xmark & \centering \cmark  &  {\hspace {3em}\xmark}  \\
         Imply Loss \citep{awasthi2020learning}  & \centering \xmark  & \centering \xmark & \centering \xmark  & \centering  \cmark & {\hspace {3em}\xmark}   \\
         Matlab  & \centering \cmark  & \centering \xmark & \centering \xmark &  \centering \xmark & {\hspace {3em}\xmark}     \\
         SkWeak \cite{lison-etal-2021-skweak}  & \centering \cmark  & \centering \cmark & \centering \cmark &  \centering \xmark & {\hspace {3em}\xmark}     \\
         Wrench \cite{zhang2021wrench}  & \centering \cmark  & \centering \xmark & \centering \cmark &  \centering \cmark & {\hspace {3em}\xmark}     \\
         \spear  & \centering \cmark  & \centering \cmark & \centering \cmark & \centering \cmark & {\hspace {3.2em}\cmark }  \\
         \hline
    \end{tabular}
    \caption{Comparison of \spear\ against available packages.
    \label{tab:comparisonTable}}
    \end{center}
\end{table*}

\section{Designing and Applying LFs}
User interacts with the library by designing labeling functions. Similar to \citet{ratner2017snorkel}, labeling functions are python functions which take a candidate as an input and either associates class label or abstains. However, continuous LFs returns a continuous score in addition to the class label. These continuous LFs are more
natural to program and lead to improved recall \cite{oishik}.
\subsection{Designing LFs}
\spear\ uses a \texttt{@labeling\_function()} decorator to define a labeling function. Each LF, when applied on an instance, can either return a class label or not return anything, \textit{i.e.} abstain. The LF decorator has an additional argument that accepts a list of preprocessors. Each preprocessor can be either  declared  as  a  pre-defined  function  or  can employ external libraries. The pre-processor transforms the data point before applying the labeling function.

\begin{lstlisting}
@labeling_function(cont_scorer, resources, preprocessors, label)
def CLF1(x,**kwargs):
  return label if kwargs["continuous_score"] >= threshold else ABSTAIN
\end{lstlisting}

The LF can express pattern matching rules in the form of heuristics, distant supervision by using external knowledge bases and other data resources to label datapoints. LFs on SMS dataset can be seen in the example notebook \href{https://github.com/decile-team/spear/blob/main/notebooks/SMS_SPAM/sms_labeling.ipynb}{here}.

\paragraph{Continuous LFs:} 
In the discrete LFs, users construct heuristic patterns based on dictionary lookups or thresholded distance for the classification tasks. However, the keywords in hand-crafted dictionaries might be incomplete. \citet{oishik} proposed a comprehensive alternative that design continuous valued LFs that return scores derived from soft match between words in the sentence and
the dictionary.

\spear\ provides the facility to declare continuous LFs, each of which returns the associated label along with a confidence score using the \texttt{@continuous\_scorer} decorator. 
The continuous score can be accessed in the LF definition through the keyword argument \texttt{continuous\_score}. As evident from Table~\ref{tab:comparisonTable}, no other existing package provisions for both semi-supervised aggregation and subset selection modules.
\begin{lstlisting}
@continuous_scorer()
def similarity(sentence,**kwargs):
  word_vecs = featurizer(sentence)
  keyword_vecs = featurizer(kwargs["keywords"])
  return similarity(word_vecs,keyword_vecs)
\end{lstlisting}

\subsection{Applying LFs}
\begin{figure*}[!h]
    \centering
    \includegraphics[width=0.8\linewidth]{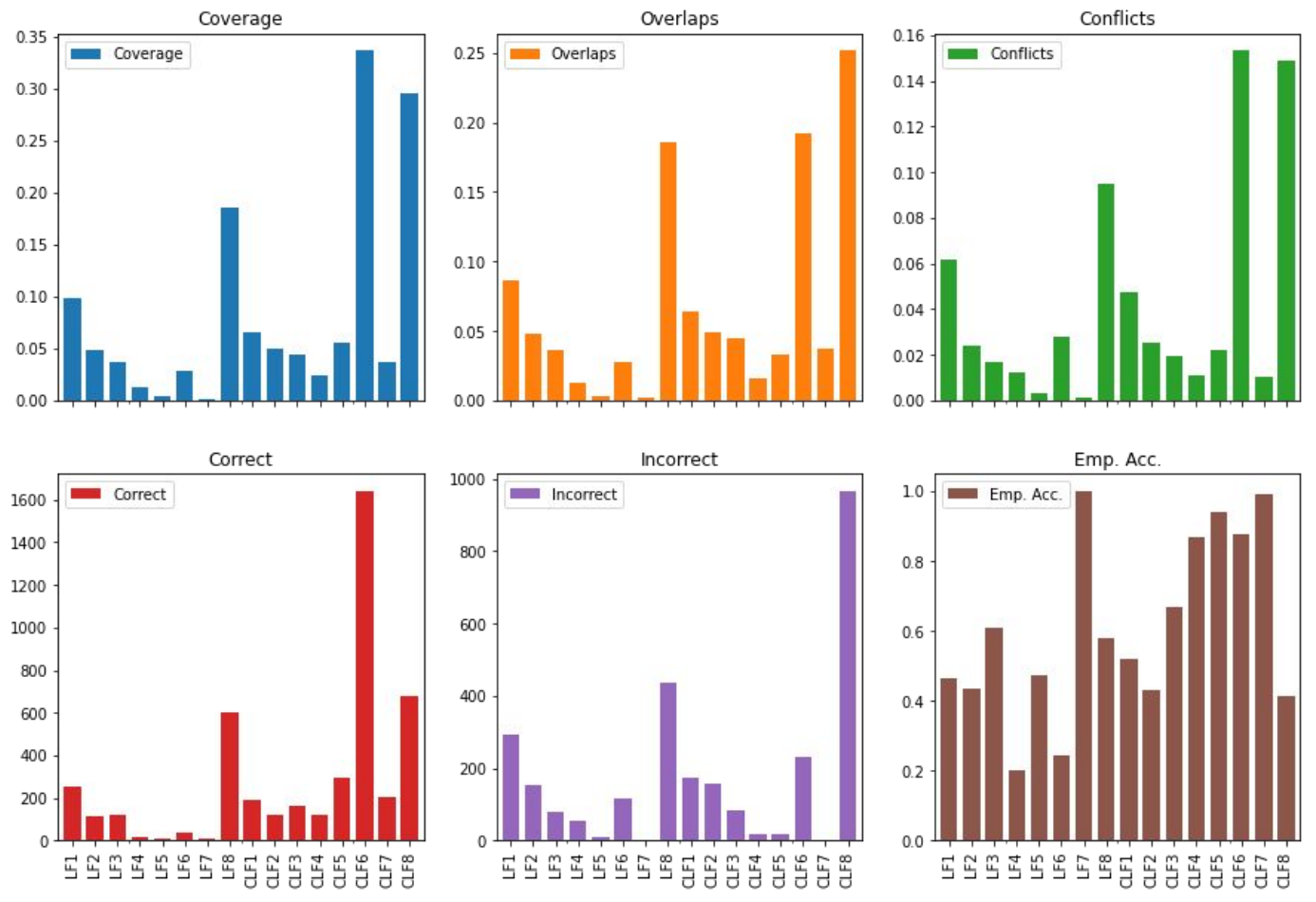}
    \caption{LF analysis on the SMS dataset presented in the form of graph visualization within the \spear\ tool. The statistics include precision, coverage, conflict and empirical accuracy for each LF.}
    \label{fig:metric}
\end{figure*}

Once LFs are defined, users can analyse labeling functions by calculating coverage, overlap, conflicts, empirical accuracy for each LF which helps to re-iterate on the process by refining new LFs. The metrics can be visualised within the \spear\ tool, either in the form of a table or graphs as shown in Figure \ref{fig:metric}.

PreLabels is the master class which encapsulates a set of LFs, the dataset to label and enum of class labels. PreLabels facilitates the process of applying the LFs on the dataset, and of analysing and refining the LF set. We provide functions to store labels assigned by LFs 
and associated meta-data such as mapping of class name to numeric class labels on the disk in the form json file(s). The pre-labeling performed using the LFs can be consolidated into labeling performed using several consensus models described in Section~\ref{sec:models}.
\begin{lstlisting}
sms_pre_labels = PreLabels(name="sms", data=X_V, gold_labels=Y_V, 
data_feats=X_feats_V, rules=rules, labels_enum=ClassLabels, num_classes=2)
\end{lstlisting}


\section{Models} \label{sec:models}
We implement several data-programming approaches in this demonstration that includes simple baselines such as fully-supervised, semi-supervised and unsupervised approaches.

\subsection{Joint Learning \cite{spear}}
    
The joint learning (JL) module implements a semi-supervised data programming paradigm that learns a joint model over LFs and features. JL has two key components, {\em viz.}, feature model (fm) and graphical model (gm) and their sum is used as a training objective.
During training, the JL requires labeled ($\Lcal$), validation ($\Vcal$), test ($\Tcal$) sets consisting of true labels and an unlabeled ($\Ucal$) set whose true labels are to be inferred. The model API closely follows that of \textit{scikit-learn}~\citep{pedregosa2011scikit} to make the package easily accessible to the machine learning audience. The primary functions are: (1) \texttt{fit\_and\_predict\_proba}, which trains using the prelabels assigned by LFs and true labels of $\Lcal$ data and predicts the probabilities of labels for each instance of $\Ucal$ data 
(2) \texttt{fit\_and\_predict}, similar to the previous one but which predicts labels of $\Ucal$ using maximum posterior probabilities 
(3) \texttt{predict\_(fm/gm)\_proba}, predicts the probabilities, using feature model(fm)/graphical model(gm) 
(4) \texttt{predict\_(fm/gm)}, predicts labels using fm/gm based on learned parameters.
We also provide functions \texttt{save} or \texttt{load\_params} to save or load the trained parameters.

As another unique feature ({\em c.f.} Table~\ref{tab:comparisonTable}), our library supports a \textit{subset-selection framework} that makes the best use of human-annotation efforts. The $\Lcal$ set can be chosen using submodular functions such as facility location, max cover, { \em etc.} We utilise the submodlib\footnote{\url{https://github.com/decile-team/submodlib}} library for the subset selection algorithms.
The function alternatives for subset selection are \texttt{rand\_subset, unsup\_subset, sup\_subset\_indices, sup\_subset\_save\_files}. 
\subsection{Only-$\Lcal$}
In this, the classifier $P(y|\bfx)$ is trained only on the labeled data. Following \citet{spear}, we provide facility to  use either Logistic Regression or a 2-layered neural network. Our package is flexible to allow other architectures to be plugged-in as well.

\begin{figure*}[t]
    \centering
    \includegraphics[width=.8\linewidth]{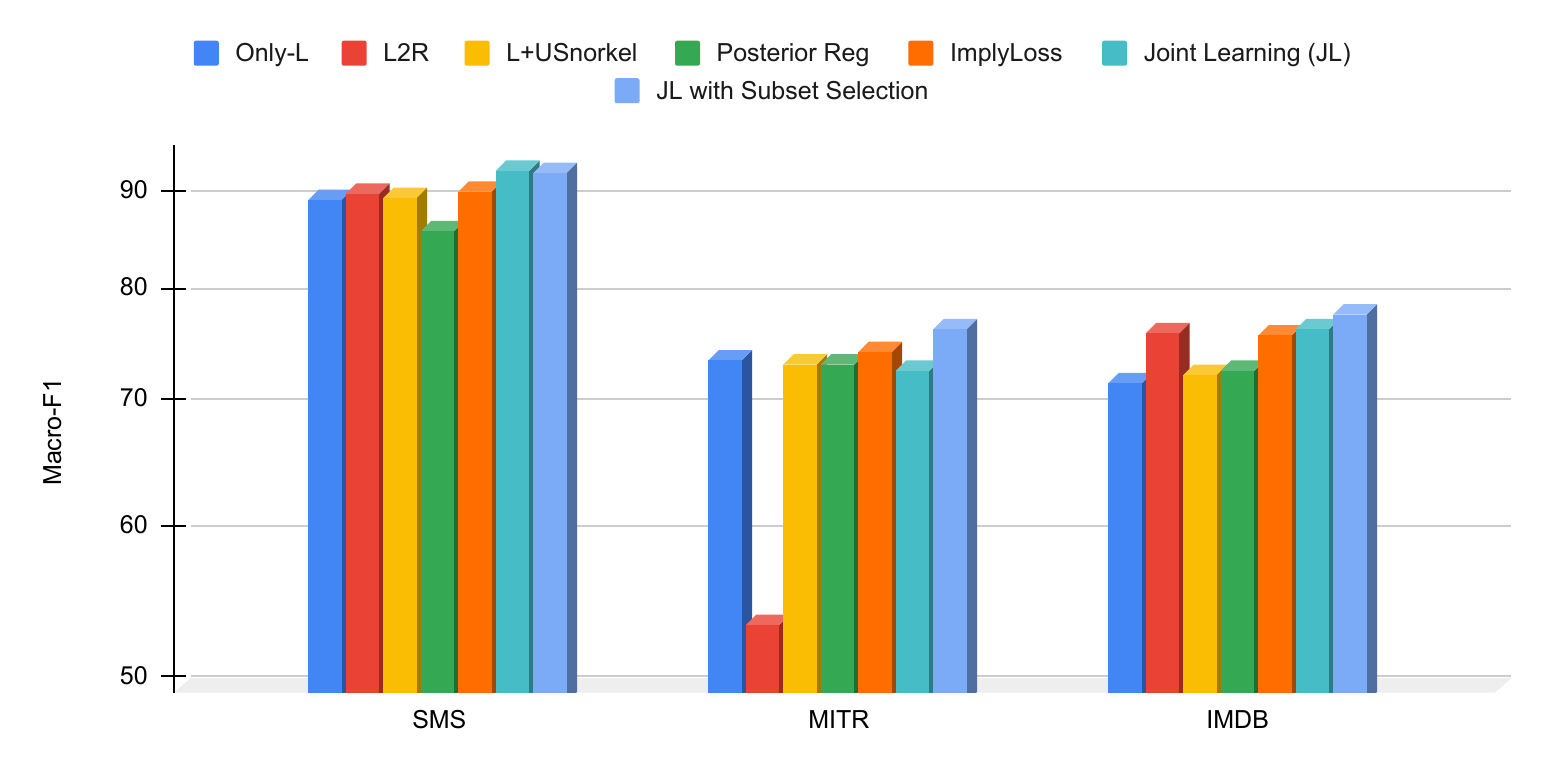}
    \caption{Experiments on SMS, IMDB and MIT-R dataset and comparison with various approaches. We use JL combined with supervised subset selection for obtaining numbers. }
    \label{fig:chart}
\end{figure*}

\subsection{\cage~\citep{oishik}}
This accepts both continuous and discrete LFs. Further, each LF has an associated quality guide component, that refers to the fraction of times the LF predicts the correct label; this stabilises training in absence of $\Vcal$ set. In our package, \cage\ accepts $\Ucal$ and $\Tcal$ sets during training. \cage{} has member functions similar to (except there are no fm or gm variants to \texttt{predict\_proba}, \texttt{predict} functions in Cage) JL module, with different arguments, serving the same purpose. It should be noted that this model doesn't need labeled($\Lcal$) or validation($\Vcal$) data.  

\subsection{Learning to Reweight (L2R) \citep{l2r}}
This method is an online meta-learning approach for reweighting training examples with a mix of $\Ucal$ and $\Lcal$. It leverages validation set to determine and adaptively assigns importance weights to examples based on the gradient direction. This does not employ additional parameters to weigh or denoise individual rules.

\subsection{$\mathbf{\Lcal+\Ucal_{Snorkel}}$~\citep{ratner2017snorkel}} 

This method trains a supervised classifier on $\Lcal$ set and Snorkel's generative model on $\Ucal$ set. Snorkel is a generative model that models class probabilities based on discrete LFs for consensus on the noisy and conflicting labels. It assigns a linear weight to each rule based on an agreement objective and label examples in $\Ucal$.

\subsection{Posterior Regularization (PR)~\citep{Hu2016Harnessing}}
This is a method that enables to simultaneously learn from $\Lcal$ and logic rules by jointly learning a rule and feature network in a teacher-student setup. The student network learns parameter $\theta$ using the $\Lcal$ set and teacher networks attempts to imitates the student network in a joint learning manner. The teacher network encodes logic rules as a regularization term in the overall loss objective.

\subsection{Imply Loss~\citep{awasthi2020learning}}
This approach uses additional information in the form of labeled rule exemplars and trains with a denoised rule-label loss.  They leverage both rules and labeled data by mapping each rule with exemplars of correct firings (i.e., instantiations) of that rule. Their joint training algorithms
denoise over-generalized rules and train a classification model. It has two main components:
\begin{enumerate}
\item {Rule Network}: It learns to predict whether a given rule has overgeneralized on a given sample using latent coverage variables. 
\item  Classification Network: It is trained on $\Lcal$ and $\Ucal$ to predict the output label and maximize the accuracy on unseen test instances using a soft implication loss. 
\end{enumerate}

This module contains the following primary classes:
\begin{enumerate}
\item  DataFeeder - It will essentially take all the parameters as input and create a data feeder class with all these parameters as its attributes. 
\item HighLevelSupervisionNetwork (HLS) - It will take the 2 networks, the mode or the approach that needs to be used to train the model, the required parameters, the directory storing model checkpoints at different instances and the instances and labels from the labeled dataset ($\Lcal$) and create an object named "hls".
\end{enumerate}
HLS object will have many member functions of which the 2 significant are:\\
(a) \textbf{hls.train}: This function, when called with the required mode, will train the 2 network attributes of the object.\\
(b) \textbf{hls.test}: It supports 3 types of testing:\\
\indent  (i) test\_w: this will test the rule network and the related model of the object.\\
\indent  (ii) test\_f: this will test the classification network and the related model of the object.\\
\indent  (iii) test\_all: this will test both the networks and models of the class.



\begin{figure*}[t]
    \centering
    \includegraphics[width=0.85\linewidth]{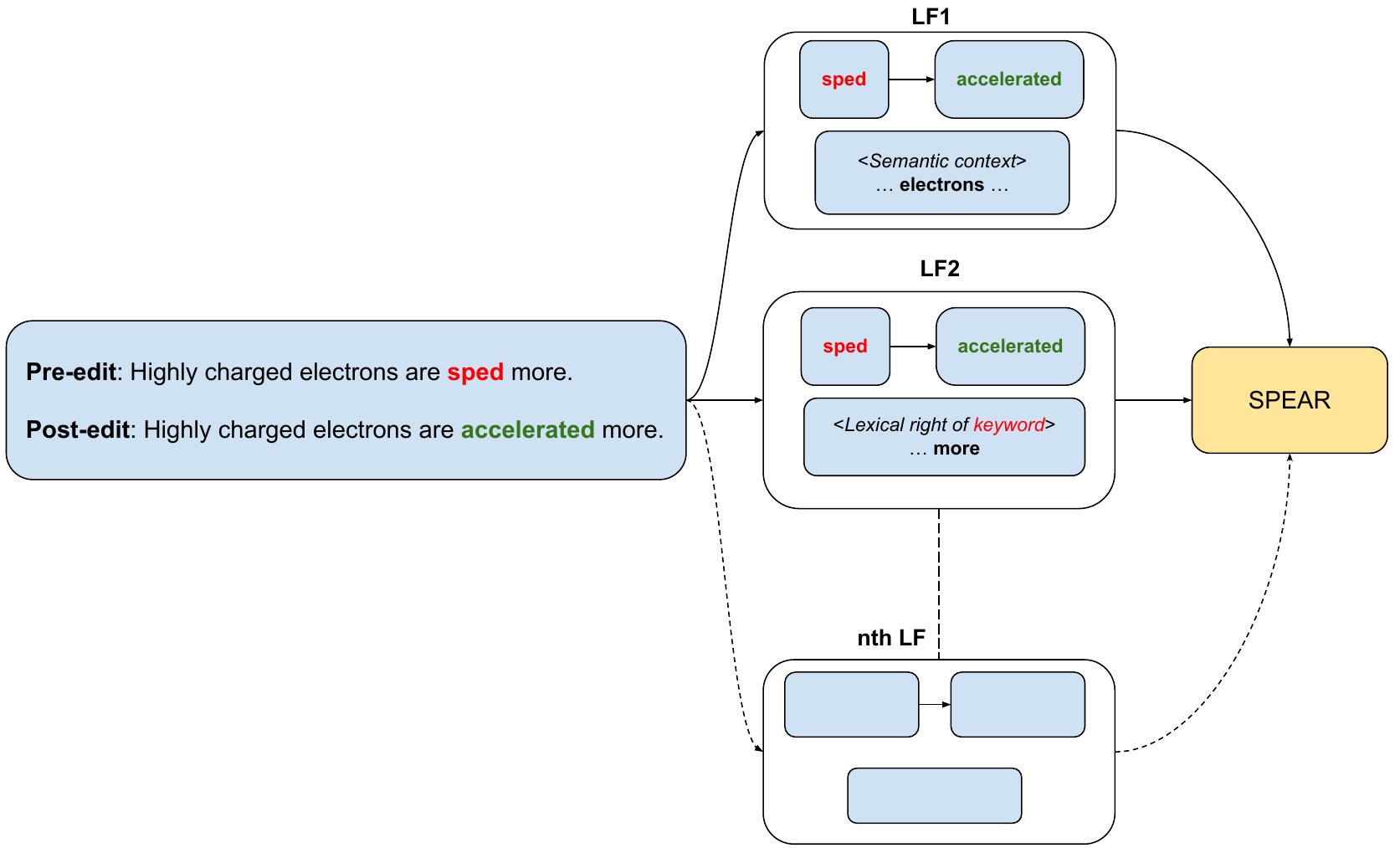}
    \caption{Mutiple LFs generated from post-editor edits based on semantic and lexical features while editing science (domain-specific) document in English.}
    \label{fig:framework}
\end{figure*}
\section{Experiments}

We prepared \href{https://github.com/decile-team/spear/tree/main/notebooks/}{jupyter tutorial notebooks} for two standard text classification datasets, namely SMS, YouTube and TREC. We took LFs on these datasets from \citet{awasthi2020learning} and train using approaches implemented in this paper.  Figure \ref{fig:chart} shows performance of various approaches implemented using our package on additional two datasets, MIT-R and IMDB. We can integrate image classification tasks by defining appropriate feature extraction module and rules.

\section {Use Cases}

\spear\ is employed in project UDAAN\footnote{\url{https://www.udaanproject.org/}} for reducing post editing efforts. UDAAN \cite{udaan} is a post-editing workbench to translate content in native languages. Based on the post editor's patterns of changes to the target language document, candidate labeling functions are generated (based on a combination of heuristics and linguistic patterns) by the UDAAN workbench ({\em c.f.} Figure~\ref{fig:framework} for examples of LFs). Based on these LFs, \spear\ gets invoked on a combination of the edited ({\em i.e.}, labeled) data and the not yet edited ({\em i.e.}, unlabeled) data to present consolidated edits to the post-editor. This use case has been presented in the flow chart in Figure~\ref{fig:framework}. We present the appropriate incorporation of \spear\ into the post-editing environment of an ecosystem such as for translation (UDAAN) or even for Optical Character Recognition\footnote{\url{https://www.cse.iitb.ac.in/~ocr/}} or Automatic Speech Recognition (ASR). 

As a part of COVID-19 third wave preparedness, \spear\ was used by the Municipal Corporation of Greater Mumbai (MCGM)’s Health Ward\footnote{\url{https://colab.research.google.com/drive/1tNUObqSDypUos7YNvnqvemALlkrrsB0z}} for predicting the COVID-19 status of patients to help in preliminary diagnosis. 

\subsection{Demonstration Case}
For the demonstration use case, apart from the use cases outlined in the previous section, we can choose a text classification dataset and form regex or continuous rules by observing few data points. Once LFs are developed, we can easily compare the LFs with any of the semi- and un-supervised algorithms present in the package. 

\section{Conclusion and Future Work}
\spear\ is a unified package for semi-supervised data programming that  quickly annotates training data and train machine learning models. 
It eases the use of developing LFs and label aggregation approaches. This allows for better reproducibility, benchmarking and easier ML development in low-resource settings such as textual post-editing. 
Presently, we are integrating automatic LF induction approaches such as Snuba \cite{varma2018snuba} that uses a small labeled set to induce LFs automatically. This will significantly increase the scope of datasets without needing human intervention in designing LFs. The package is written in Python3 and open-sourced with a MIT License\footnote{\url{https://opensource.org/licenses/MIT}} open for community contribution.

\section{Acknowledgements}
We thank anonymous reviewers for providing constructive feedback. Ayush Maheshwari is supported by a Fellowship from Ekal Foundation (www.ekal.org). Ganesh Ramakrishnan is grateful to IBM Research, India (specifically the IBM AI Horizon Networks - IIT Bombay initiative) as well
as the IIT Bombay Institute Chair Professorship for
their support and sponsorship.





\bibliographystyle{acl_natbib}
\bibliography{refs}

\begin{thebibliography}{15}
\expandafter\ifx\csname natexlab\endcsname\relax\def\natexlab#1{#1}\fi

\bibitem[{Awasthi et~al.(2020)Awasthi, Ghosh, Goyal, and
  Sarawagi}]{awasthi2020learning}
Abhijeet Awasthi, Sabyasachi Ghosh, Rasna Goyal, and Sunita Sarawagi. 2020.
\newblock \href {https://openreview.net/forum?id=SkeuexBtDr} {Learning from
  rules generalizing labeled exemplars}.
\newblock In \emph{8th International Conference on Learning Representations,
  {ICLR} 2020, Addis Ababa, Ethiopia, April 26-30, 2020}. OpenReview.net.

\bibitem[{Bach et~al.(2019)Bach, Rodriguez, Liu, Luo, Shao, Xia, Sen, Ratner,
  Hancock, Alborzi et~al.}]{bach2019snorkel}
Stephen~H Bach, Daniel Rodriguez, Yintao Liu, Chong Luo, Haidong Shao,
  Cassandra Xia, Souvik Sen, Alex Ratner, Braden Hancock, Houman Alborzi,
  et~al. 2019.
\newblock Snorkel drybell: A case study in deploying weak supervision at
  industrial scale.
\newblock In \emph{Proceedings of the 2019 International Conference on
  Management of Data}, pages 362--375.

\bibitem[{Chatterjee et~al.(2020)Chatterjee, Ramakrishnan, and
  Sarawagi}]{oishik}
Oishik Chatterjee, Ganesh Ramakrishnan, and Sunita Sarawagi. 2020.
\newblock Robust data programming with precision-guided labeling functions.
\newblock In \emph{AAAI}.

\bibitem[{Hu et~al.(2016)Hu, Ma, Liu, Hovy, and Xing}]{Hu2016Harnessing}
Zhiting Hu, Xuezhe Ma, Zhengzhong Liu, Eduard Hovy, and Eric Xing. 2016.
\newblock Harnessing deep neural networks with logic rules.
\newblock \emph{arXiv preprint arXiv:1603.06318}.

\bibitem[{Lison et~al.(2021)Lison, Barnes, and Hubin}]{lison-etal-2021-skweak}
Pierre Lison, Jeremy Barnes, and Aliaksandr Hubin. 2021.
\newblock \href {https://doi.org/10.18653/v1/2021.acl-demo.40} {skweak: Weak
  supervision made easy for {NLP}}.
\newblock In \emph{Proceedings of the 59th Annual Meeting of the Association
  for Computational Linguistics and the 11th International Joint Conference on
  Natural Language Processing: System Demonstrations}, pages 337--346, Online.
  Association for Computational Linguistics.

\bibitem[{Maheshwari et~al.(2021)Maheshwari, Chatterjee, Killamsetty, Iyer, and
  Ramakrishnan}]{spear}
Ayush Maheshwari, Oishik Chatterjee, KrishnaTeja Killamsetty, Rishabh~K. Iyer,
  and Ganesh Ramakrishnan. 2021.
\newblock \href {http://arxiv.org/abs/2008.09887} {Data programming using
  semi-supervision and subset selection}.
\newblock In \emph{Proceedings of the 59th Annual Meeting of the Association
  for Computational Linguistics}.

\bibitem[{Maheshwari et~al.(2022{\natexlab{a}})Maheshwari, Killamsetty,
  Ramakrishnan, Iyer, Danilevsky, and Popa}]{maheshwari2022learning}
Ayush Maheshwari, Krishnateja Killamsetty, Ganesh Ramakrishnan, Rishabh Iyer,
  Marina Danilevsky, and Lucian Popa. 2022{\natexlab{a}}.
\newblock Learning to robustly aggregate labeling functions for semi-supervised
  data programming.
\newblock In \emph{Findings of the Association for Computational Linguistics:
  ACL 2022}, pages 1188--1202.

\bibitem[{Maheshwari et~al.(2022{\natexlab{b}})Maheshwari, Ravindran,
  Subramanian, Jalan, and Ramakrishnan}]{udaan}
Ayush Maheshwari, Ajay Ravindran, Venkatapathy Subramanian, Akshay Jalan, and
  Ganesh Ramakrishnan. 2022{\natexlab{b}}.
\newblock \href {https://doi.org/10.48550/ARXIV.2203.01644} {Udaan -- machine
  learning based post-editing tool for document translation}.

\bibitem[{Pedregosa et~al.(2011)Pedregosa, Varoquaux, Gramfort, Michel,
  Thirion, Grisel, Blondel, Prettenhofer, Weiss, Dubourg
  et~al.}]{pedregosa2011scikit}
Fabian Pedregosa, Ga{\"e}l Varoquaux, Alexandre Gramfort, Vincent Michel,
  Bertrand Thirion, Olivier Grisel, Mathieu Blondel, Peter Prettenhofer, Ron
  Weiss, Vincent Dubourg, et~al. 2011.
\newblock Scikit-learn: Machine learning in python.
\newblock \emph{the Journal of machine Learning research}, 12:2825--2830.

\bibitem[{Ratner et~al.(2017)Ratner, Bach, Ehrenberg, and
  R{\'e}}]{ratner2017snorkel}
Alexander~J Ratner, Stephen~H Bach, Henry~R Ehrenberg, and Chris R{\'e}. 2017.
\newblock Snorkel: Fast training set generation for information extraction.
\newblock In \emph{Proceedings of the 2017 ACM international conference on
  management of data}, pages 1683--1686.

\bibitem[{Ren et~al.(2018)Ren, Zeng, Yang, and Urtasun}]{l2r}
Mengye Ren, Wenyuan Zeng, Bin Yang, and Raquel Urtasun. 2018.
\newblock Learning to reweight examples for robust deep learning.
\newblock In \emph{International Conference on Machine Learning}, pages
  4334--4343.

\bibitem[{Sahay et~al.(2021)Sahay, Nasery, Maheshwari, Ramakrishnan, and
  Iyer}]{sahay2021rule}
Atul Sahay, Anshul Nasery, Ayush Maheshwari, Ganesh Ramakrishnan, and Rishabh
  Iyer. 2021.
\newblock Rule augmented unsupervised constituency parsing.
\newblock In \emph{Findings of the Association for Computational Linguistics:
  ACL-IJCNLP 2021}, pages 4923--4932.

\bibitem[{Sen et~al.(2020)Sen, Danilevsky, Li, Brahma, Boehm, Chiticariu, and
  Krishnamurthy}]{sen2020learning}
Prithviraj Sen, Marina Danilevsky, Yunyao Li, Siddhartha Brahma, Matthias
  Boehm, Laura Chiticariu, and Rajasekar Krishnamurthy. 2020.
\newblock Learning explainable linguistic expressions with neural inductive
  logic programming for sentence classification.
\newblock In \emph{Proceedings of the 2020 Conference on Empirical Methods in
  Natural Language Processing (EMNLP)}, pages 4211--4221.

\bibitem[{Varma and R{\'e}(2018)}]{varma2018snuba}
Paroma Varma and Christopher R{\'e}. 2018.
\newblock Snuba: automating weak supervision to label training data.
\newblock In \emph{Proceedings of the VLDB Endowment. International Conference
  on Very Large Data Bases}, volume~12, page 223. NIH Public Access.

\bibitem[{Zhang et~al.(2021)Zhang, Yu, Li, Wang, Yang, Yang, and
  Ratner}]{zhang2021wrench}
Jieyu Zhang, Yue Yu, Yinghao Li, Yujing Wang, Yaming Yang, Mao Yang, and
  Alexander Ratner. 2021.
\newblock Wrench: A comprehensive benchmark for weak supervision.
\newblock In \emph{Thirty-fifth Conference on Neural Information Processing
  Systems Datasets and Benchmarks Track (Round 2)}.

\end{thebibliography}

\end{document}